\documentclass[letterpaper, 10 pt, conference]{ieeeconf}  
\usepackage{graphicx}
\usepackage{amsmath}
\usepackage{capt-of}
\usepackage{booktabs}
\usepackage{array}
\usepackage{amssymb}
\usepackage{hyperref}
\makeatletter
\let\athlete@makecaption\@makecaption
\long\def\@makecaption#1#2{%
    \ifx\@captype\@IEEEtablestring
            \par\vskip 3pt
            {\centering\footnotesize #1.~#2\par}%
            \vskip 2pt
    \else
        \athlete@makecaption{#1}{#2}%
    \fi}
\makeatother

\IEEEoverridecommandlockouts                              

\title{\LARGE \bf
ATHLETE: Learning Reactive Humanoid Tennis via Trajectory-Guided Motion Matching
}

\author{
Zhihao Tian$^{1,*}$,
Zifan Wang$^{1,*}$,
Enzheng Zhu$^{1}$,
Xun Yang$^{1}$,
Jialong Qin$^{1}$,
Mengbo Wang$^{1}$,
and Junwei Liang$^{1,\dagger}$%
\thanks{$^{*}$Equal contribution.}%
\thanks{$^{\dagger}$Corresponding author: Junwei Liang.
Email: {\tt\small junweiliang1114@gmail.com}.}%
\thanks{$^{1}$The Hong Kong University of Science and Technology
(Guangzhou), Guangzhou, China.}%
}

\IEEEaftertitletext{%
    \begin{minipage}{\textwidth}
        \centering
        \includegraphics[width=0.9\textwidth]{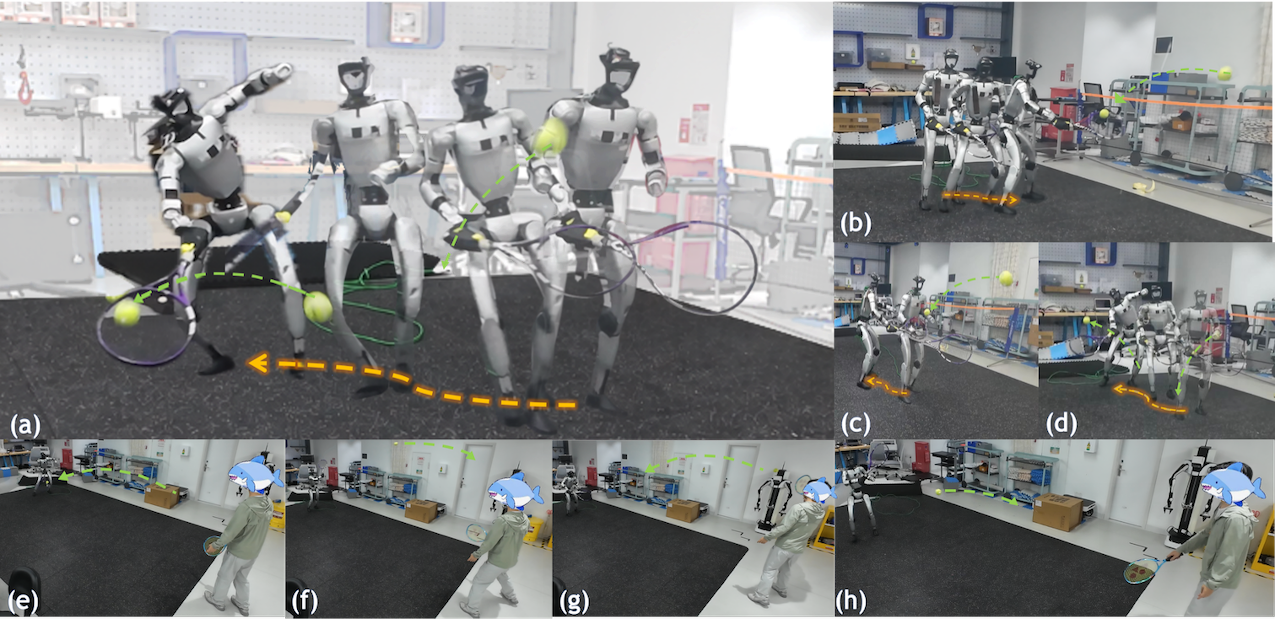}
        \captionof{figure}{Real-world humanoid tennis with ATHLETE, which enables humanoid robots to learn tennis skills from generic locomotion data and ordinary tennis videos. 
        The ATHLETE policy learns to respond to incoming balls through whole-body locomotion and striking motions, 
        without requiring external reference-motion inputs or an explicit motion planner at deployment. 
        As shown in the figures, the policy responds to different incoming-ball trajectories by coordinating whole-body locomotion to approach the ball and automatically using a forehand or backhand swing to strike it.}
        \label{fig:athlete_teaser}
    \end{minipage}\vspace{0.5\baselineskip}
}

\begin{document}

\maketitle
\thispagestyle{empty}
\pagestyle{empty}

\begin{abstract}

Humanoid tennis requires reactive whole-body coordination under rapidly changing ball dynamics, yet collecting task-specific motion-capture demonstrations that cover diverse approach states, contact locations, and timing conditions is costly. Inspired by how humans acquire new skills by combining reusable motor experience with task-specific demonstrations, we present ATHLETE, a framework for learning reactive humanoid tennis from generic locomotion data and ordinary tennis videos. ATHLETE combines LAFAN1 locomotion with video-derived striking motions and uses Trajectory-Guided Motion Matching (TGMM) to construct diverse task-conditioned whole-body references. These references are transformed into a reactive policy through physics-based teacher--student learning with online DAgger distillation, followed by PPO fine-tuning for the tennis task. The resulting policy responds directly to robot and ball observations without requiring reference-motion inputs at deployment. On a real humanoid robot, ATHLETE achieves forehand and backhand hit rates of $90.00\%$ and $86.67\%$, respectively, and enables multi-shot rallies with human players. These results demonstrate that ATHLETE can learn dynamic humanoid tennis skills from reusable locomotion data and video demonstrations.  Result videos and code can be found at \url{https://anonymous.4open.science/w/ATHLETE2027/}.

\end{abstract}
\vspace{-0.4cm}

\section{INTRODUCTION}

Humanoid tennis requires rapid locomotion, precisely timed whole-body strikes, and continuous adaptation to incoming balls.
A central challenge is data coverage: a successful policy must handle diverse approach states, contact locations, and arrival times, while collecting task-specific motion-capture demonstrations for all such variations is costly.

Existing works such as LATENT~\cite{zhang2026latent} and AdaPT~\cite{huang2026adapt} demonstrate that human motion data can provide strong priors for humanoid tennis.
However, collecting or reconstructing task-specific tennis motions that sufficiently cover the combinations of locomotion, contact position, and timing remains difficult.
At the same time, public motion datasets such as LAFAN1~\cite{harvey2020robust} and AMASS~\cite{mahmood2019amass} already contain abundant reusable human locomotion, while ordinary tennis videos provide task-specific striking motions.
This raises a natural question: \emph{can dynamic humanoid tennis be learned by composing reusable locomotion experience with only sparse task-specific striking demonstrations?}

We present ATHLETE to answer this question.
Our key idea is to treat locomotion and striking as complementary motion sources rather than requiring complete tennis demonstrations.
ATHLETE uses Trajectory-Guided Motion Matching (TGMM) to compose generic locomotion with video-derived strikes under task, contact, and transition constraints, producing diverse whole-body approach-and-strike references.
These references are then converted through physics-based teacher--student learning into a reactive policy that no longer requires reference-motion input, online motion retrieval, or explicit motion planning at deployment.
The resulting controller directly responds to the robot state, ball observations, and task target.

Our contributions are threefold:
\begin{itemize}

    \item \textbf{Task-conditioned motion composition from heterogeneous motion sources.}
    We introduce TGMM to combine abundant generic locomotion with sparse video-derived striking motions.
    By explicitly enforcing approach, contact, and transition constraints, the method generates diverse whole-body tennis references without collecting dedicated tennis motion-capture demonstrations.

    \item \textbf{From composed references to reference-free reactive control.}
    We develop a deadline-adaptive teacher--student learning framework that transfers reference-motion and contact-timing supervision into a deployable reactive policy.
    The final controller operates without reference motions, online motion matching, or an explicit motion planner.

    \item \textbf{Real-world humanoid tennis from low-cost motion sources.}
    We validate ATHLETE on a humanoid robot, achieving reactive forehand and backhand returns and multi-shot human--robot rallies, demonstrating that reusable locomotion data and sparse visual demonstrations are sufficient to support dynamic whole-body tennis skills (Fig.~\ref{fig:athlete_teaser}).
\end{itemize}

\begin{figure}[t]
    \vspace*{2mm}
    \centering
    \includegraphics[width=\columnwidth]{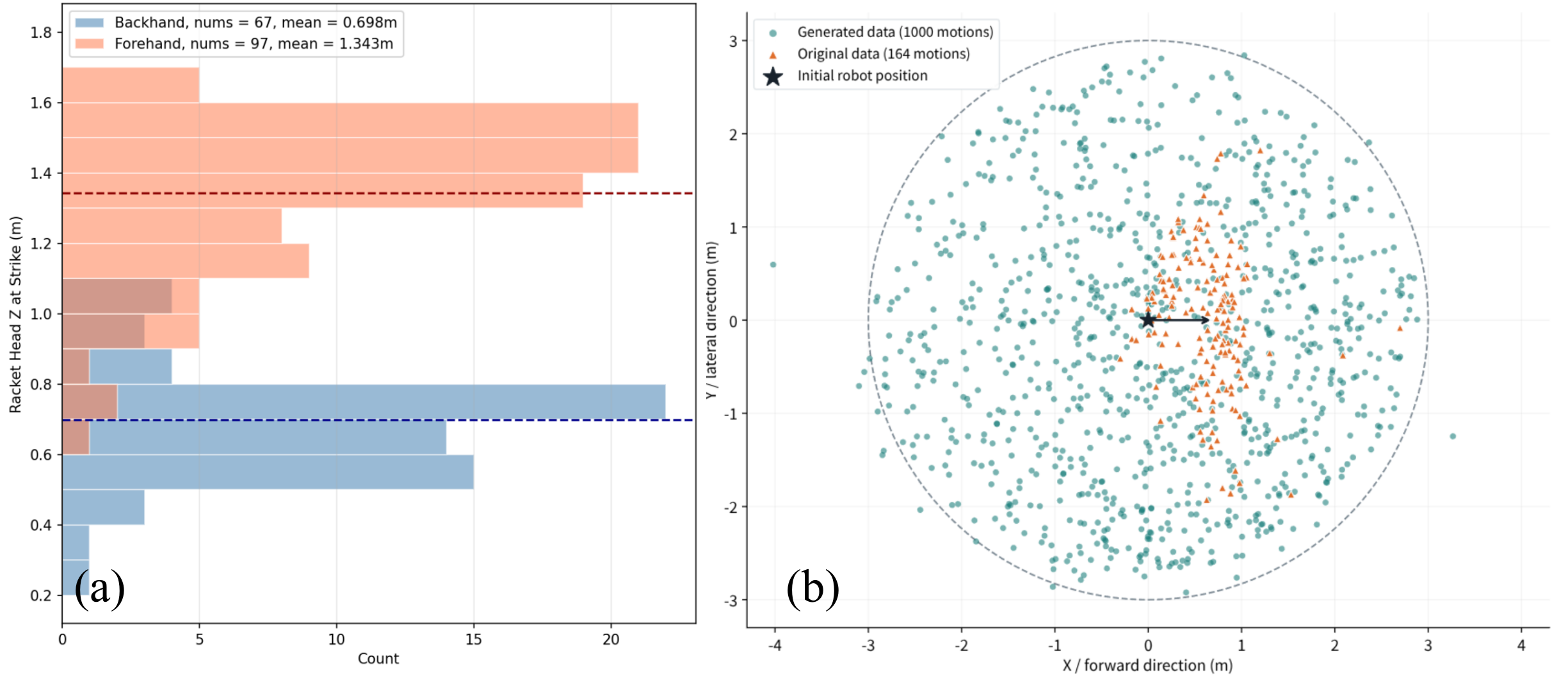}
    \caption{Strike-point height and XY distribution of source skills and generated references.
    (a) strike height distribution of source tennis references.
    (b) XY distribution of strike points in local frames.}
    \label{fig:strike_distribution}
\end{figure}
\setlength{\textfloatsep}{3pt}


\vspace*{-2mm}
\section{RELATED WORKS}

\subsection{Data Generation for Humanoid Robots}
Motion matching retrieves and blends database motions using pose and trajectory features~\cite{clavet2016motionmatching}, while Learned Motion Matching approximates this pipeline with neural networks to reduce memory usage~\cite{holden2020learned}.
Perceptive Humanoid Parkour~\cite{wu2026perceptiveparkour} combines sparse atomic skill demonstrations with locomotion to generate long-horizon references with diverse approach states.

Generative models synthesize motion from history, text, or kinematic constraints~\cite{tevet2023human,rempeKimodo2026,ling2020character,zhao2026ardy}; humanoid extensions also condition references on velocity commands, terrain, or multimodal inputs~\cite{zhang2025gmp,zhang2026wholebodylocomotion,xie2026textop,huang2026omg}.
ATHLETE uses optimized approach trajectories to compose locomotion and strikes under contact and transition constraints, followed by physics simulated tracking.

\subsection{Humanoid Whole-Body Control}
Reference-tracking methods learn whole-body skills from human motions through reference adaptation, dynamics alignment, and large-scale training~\cite{liao2025beyondmimic,he2025asap,li2025amo,luo2026sonic}.
Perception-conditioned controllers adapt these skills to terrain and obstacles~\cite{wu2026perceptiveparkour,wang2026perceptivebfm}.
Interaction-aware retargeting and imitation incorporate object and scene constraints to support whole-body interaction~\cite{yang2025omniretarget,weng2025hdmi,wang2026humanx}.

For tennis, LATENT~\cite{zhang2026latent} learns a correctable latent action space from imperfect motion primitives and composes skills through a high-level policy.
AdaPT~\cite{huang2026adapt} combines video-derived motions with hierarchical planning and speed-adaptive tracking.
TaskNPoint~\cite{werner2026tasknpoint} organizes demonstration-guided learning around critical interaction windows and randomized task targets.
ATHLETE adopts its motion-tracking and goal-conditioned striking rewards, adds deadline-conditioned phase adaptation, and distills the resulting skills into a single reactive policy.
\vspace{-0.2cm}
\section{Model-Based Reference Motion Generation}
\label{sec:data_processing}

\begin{figure*}[!t]
    \vspace*{2mm}
    \centering
    \includegraphics[width=0.9\textwidth]{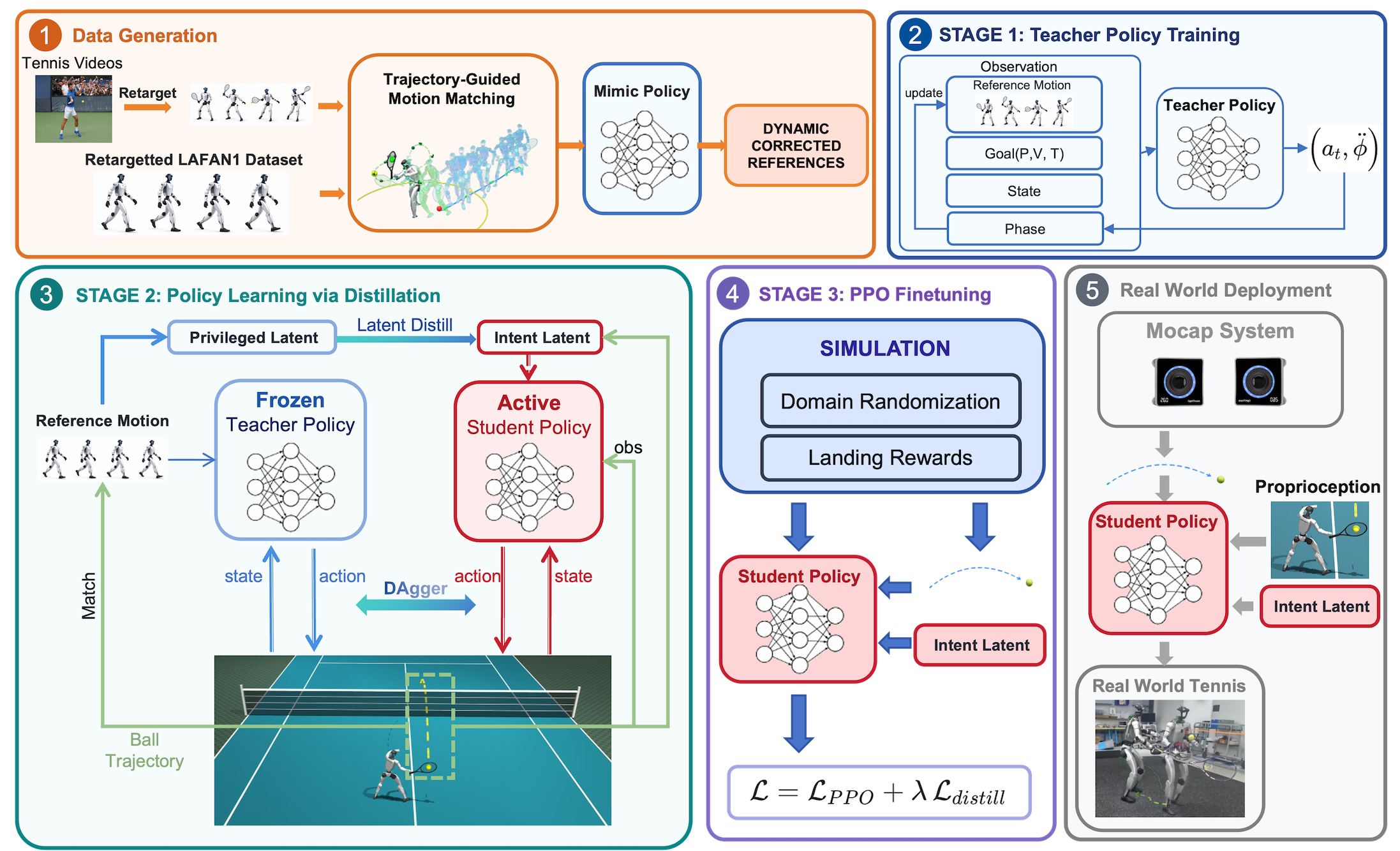}
    \caption{Overview of the ATHLETE framework.
    (1) We construct reference motions from tennis videos and a subset of the LAFAN1 dataset using TGMM, followed by mimic-policy rollouts for dynamic correction.
    (2) We train a time-adaptive teacher policy to expand the range of incoming ball trajectories that can be matched to the generated references.
    (3) We distill the teacher into a reactive student policy through online DAgger and privileged representation alignment.
    (4) We fine-tune the student using PPO with task-specific rewards and domain randomization to improve ball contact and return accuracy.
    (5) The resulting student policy is deployed on a humanoid robot to enable real-world tennis rallies with human players, using a motion-capture system for ball-state estimation.}
    \label{fig:athlete_framework}
    \vspace{-0.4cm} 
\end{figure*}

We combine locomotion sequences from the LAFAN1 dataset with video-derived striking motions through
motion matching to generate task-conditioned tennis striking references; the complete pipeline is
illustrated in Fig.~\ref{fig:athlete_framework}. We then use simulated rollouts of a Mimic policy to produce dynamically corrected references.

\subsection{Source Skills and Contact Matching}
 
Our motion library contains 16 locomotion sequences from LAFAN1~\cite{harvey2020robust}
and 67 backhand and 97 forehand clips segmented from a 10-minute video of a tennis player's
practice session.
We reconstruct human motions from these video clips using GVHMR~\cite{shen2024gvhmr}
and retarget them to the humanoid robot using GMR~\cite{araujo2025gmr}, forming the tennis
striking skill library (Fig.~\ref{fig:strike_distribution}).

\begin{figure}[!t]
    \vspace*{2mm}
    \centering
    \includegraphics[width=0.8\columnwidth]{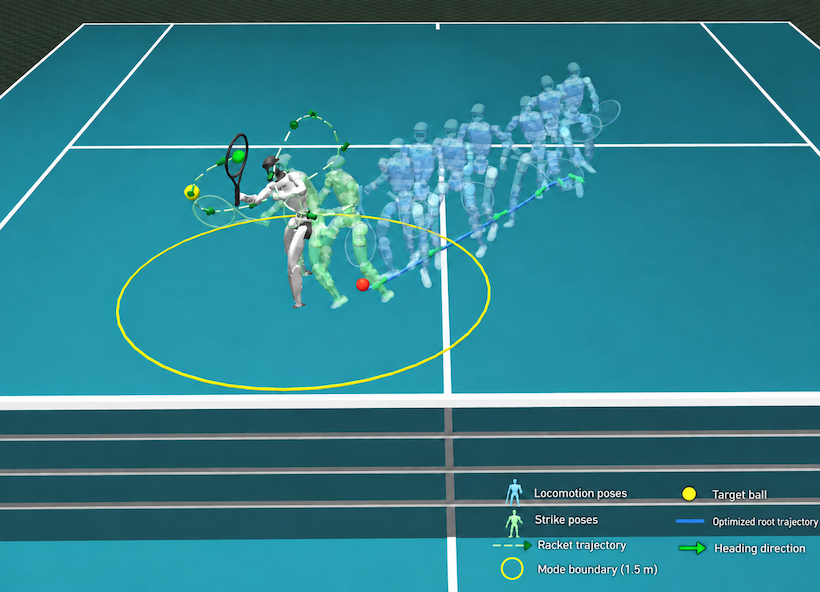}
    \caption{Strike-Motion generation using Trajectory-Guided Motion Matching (TGMM).
    The yellow ball marks the target point.
    The blue trajectories are optimized for smoothness under planar motion constraints and then converted into whole-body motion.
    The translucent blue and green robots depict the locomotion and striking phases, respectively.
    The green curve traces the trajectory of the racket's sweet spot. }
    \label{fig:model_based_motion_generation}
\end{figure}


We randomly sample a stationary target point $\mathbf{b}$ around the robot as the task target
for selecting a striking reference motion. Based on its lateral position, we first restrict the
candidates to forehand or backhand clips. We then compute the contact cost between the target and
the contact point of each candidate clip:
\begin{equation}
    E_c(i)=w_z\lvert c_{i,z}-b_z\rvert
    +w_{xy}\lVert\mathbf{c}_{i,xy}-\mathbf{b}_{xy}\rVert_2,
    \label{eq:contact_cost}
\end{equation}
where $\mathbf{c}_i$ denotes the contact point of candidate motion clip $i$. All positions in
Eq.~\eqref{eq:contact_cost} are expressed in the robot's initial root-facing coordinate frame.
We assign a larger weight to the height difference and select the clip with the lowest contact cost.

\subsection{Trajectory-Guided Motion Matching}

To enable natural humanoid locomotion that transitions into the striking clip, we use trajectory
optimization to plan a locomotion trajectory. We then use the optimized trajectory to retrieve and
match consistent whole-body locomotion from the generic motion library.

Let $d$ be the horizontal distance from the target point to the initial position, and set $d_0=1.5$\,m.
For a nearby entry position ($d\leq d_0$), the approach primarily translates while retaining its
initial heading, followed by heading alignment.
For a distant entry position, the robot turns toward the travel direction, runs forward, and
recovers the strike-entry heading.
Both modes optimize their phase durations.
We use planar state $\mathbf{z}=(\mathbf{p},\psi,v_f,v_l,\omega)$ and control
$\mathbf{u}=(a_f,a_l,a_\omega)$, with $\dot{\mathbf{p}}=R(\psi)(v_f,v_l)^\top$, $\dot\psi=\omega$,
$\dot v_f=a_f$, $\dot v_l=a_l$, and $\dot\omega=a_\omega$.

Given the current robot state $\mathbf{z}_0=\mathbf{z}_{\mathrm{init}}$, the trajectory optimizer solves:
\begin{equation}
\begin{aligned}
\min_{\{\mathbf{z}_k,\mathbf{u}_k\},\{T_m\}}\quad
J={}&w_T T+w_L L+w_u\langle\|\bar{\mathbf{u}}_k\|^2\rangle\\
&+w_{\Delta u}\Bigl\langle\Bigl\|\frac{\bar{\mathbf{u}}_{k+1}-\bar{\mathbf{u}}_k}{\Delta t_k}\Bigr\|^2\Bigr\rangle\\
&+w_eE_e+E_{\mathrm{mode}}\\
\text{s.t.}\quad
&\mathbf{z}_{k+1}=f_d(\mathbf{z}_k,\mathbf{u}_k,\Delta t_k),\\
&\|\mathbf{p}_K-\mathbf{p}_e\|\leq\epsilon_p,\quad
|\psi_K-\psi_e|\leq\epsilon_\psi,\\
&\|\mathbf{v}_K-\mathbf{v}_e\|\leq\epsilon_v,\quad
|\omega_K-\omega_e|\leq\epsilon_\omega,\\
&(\mathbf{z}_k,\mathbf{u}_k)\in\mathcal{C}_{\mathrm{mode}},\\
&\sum_mT_m\leq T_{\max}.
\end{aligned}
\label{eq:trajectory_optimization}
\end{equation}
Equation~\eqref{eq:trajectory_optimization}, in which $k$ indexes the discrete trajectory samples
and $K$ denotes the final sample, uses subscript $e$ to denote the strike-entry state.
The function $f_d$ is the discrete state-transition function obtained from the planar dynamics;
$m$ indexes the motion phases, and $T=\sum_mT_m$ is the total motion duration.

The path length $L$ is normalized by the straight-line distance from the initial position to the
strike-entry position, with the normalization distance lower-bounded by 1\,m.
The variable $\bar{\mathbf{u}}$ denotes componentwise normalized controls, and
$\langle\cdot\rangle$ denotes averaging over trajectory samples.
The terminal constraints bound position, heading, linear-velocity, and angular-velocity errors by
$\epsilon_p$, $\epsilon_\psi$, $\epsilon_v$, and $\epsilon_\omega$, respectively.
The set $\mathcal{C}_{\mathrm{mode}}$ collects mode-dependent velocity, angular-velocity, control,
and heading constraints, while the timing constraint bounds the total duration.
The costs favor short, direct approaches with smooth controls; $E_e$ penalizes terminal error, while
$E_{\mathrm{mode}}$ penalizes motion inconsistent with the selected approach mode.
We transcribe the problem using trapezoidal collocation in CasADi~\cite{andersson2019casadi} and
solve the resulting nonlinear program with IPOPT~\cite{wachter2006implementation}, then interpolate
the optimized trajectory to construct motion-matching queries.

From the optimized state sequence $\{\mathbf{z}_k\}$, we extract future planar positions, headings,
and root velocities and combine them with the current motion state to construct the motion-matching
query feature $\hat{\mathbf{f}}_{\mathrm{query}}$.
Following motion matching~\cite{clavet2016motionmatching}, we use this query to retrieve a matching
motion frame from the LAFAN1 locomotion library:
\begin{equation}
j^*=\arg\min_{j\in\mathcal{V}_L}
\|\hat{\mathbf{f}}_j-\hat{\mathbf{f}}_{\mathrm{query}}\|_W^2.
\label{eq:motion_matching}
\end{equation}
Here, $j$ indexes candidate motion frames, $\mathcal{V}_L$ is the set of eligible locomotion frames,
and $j^*$ is the match with the lowest total cost.
The normalized feature $\hat{\mathbf{f}}_j$ contains foot positions and velocities, root velocity,
racket-hand position and relative velocity, and future planar positions and headings at 0.4, 0.8,
and 1.2\,s.
The diagonal matrix $W$ weights the feature components.
By globally searching the locomotion-frame set and connecting the best-matching frames, the planned
planar trajectory is converted into a whole-body locomotion reference and connected to the striking
motion selected by contact matching (Fig.~\ref{fig:model_based_motion_generation}).

\subsection{Composition and Dynamic Correction}

At the locomotion-to-strike transition, we select the frame in the chosen strike clip whose feature
vector best matches the current pose as the entry frame, and interpolate toward it.
Horizontal root alignment and offset-decay blending reduce discontinuities. For joint positions with
initial mismatch $\Delta\mathbf{q}_0$, we apply
\begin{equation}
\mathbf{q}(t)=\mathbf{q}_{\mathrm{in}}(t)
+\left(1-\frac{t}{T_b}\right)^5\Delta\mathbf{q}_0,
\qquad 0\leq t\leq T_b.
\label{eq:offset_blending}
\end{equation}
Here, $\mathbf{q}_{\mathrm{in}}(t)$ is the strike-joint trajectory starting from the entry frame,
$\Delta\mathbf{q}_0=\mathbf{q}_{\mathrm{cur}}-\mathbf{q}_{\mathrm{in}}(0)$ is the joint offset at
the transition, and $T_b$ is the blending duration.
Root translations are blended similarly, while orientations are interpolated using quaternions.
The selected strike motion is then played until completion.

Residual discontinuities and kinematic inconsistencies can remain after blending.
Following the motion-tracking formulation of BeyondMimic~\cite{liao2025beyondmimic}, we train a
mimic policy to track the composed kinematic references and roll it out in physics simulation.
We re-collect the policy-executed trajectories as dynamics-corrected tennis motion references for
training the teacher policy in Section IV (Fig.~\ref{fig:athlete_framework}\,\textcircled{1}).

\section{Three-Stage Tennis Policy Learning}
\label{sec:policy_learning}


In this section, we use the dynamics-corrected tennis reference-motion dataset to train a
goal-conditioned and time-adaptive teacher policy.
Next, we match physically simulated ball trajectories with the reference motions used by the
teacher policy and distill a student policy that requires no reference-motion input.
Finally, in Stage III, we fine-tune the student policy using task rewards and physics domain
randomization.
\subsection{Stage I: Time-Adaptive Teacher Policy}

Fixed-rate reference playback allows the tracking policy to reach the striking state only at a
predetermined time, limiting its ability to handle balls with different speeds and flight times.
To adapt the same striking reference to different contact times, we train a time-adaptive teacher
policy (Fig.~\ref{fig:athlete_framework}\,\textcircled{2}) .

For reference motion $i$, let $\phi_t\in[0,1]$ denote the current normalized phase,
$\phi_i^{\mathrm{hit}}$ its strike phase, and $\bar T_i$ the time required to reach the strike phase
at the nominal playback rate.
We sample a target strike deadline $T_i$ at 0.1\,s intervals from
$[\bar T_i/\kappa,\bar T_i]$, where $\kappa=2$.
At time step $t$, the teacher observes the current robot state $\mathbf{s}_t$, reference pose
$\mathbf{q}(\phi_t)$, task goal $\mathbf{G}$, and remaining time $T_t^{\mathrm{rem}}$, and outputs
the joint action $\mathbf{a}_t$ and phase acceleration $\ddot\phi_t$:
\begin{equation}
(\mathbf{a}_t,\ddot\phi_t)=\pi_T\left(
\mathbf{s}_t,
\mathbf{q}(\phi_t),
\mathbf{G},
T_t^{\mathrm{rem}}
\right).
\label{eq:teacher_policy}
\end{equation}
The remaining time is $T_t^{\mathrm{rem}}=T_i-t\Delta t$.
We train the teacher with Proximal Policy Optimization (PPO)~\cite{schulman2017proximal} to strike a stationary ball at the sampled deadline.
The reward uses the motion-tracking and goal-conditioned striking terms from
TaskNPoint~\cite{werner2026tasknpoint}, augmented with a phase-timing reward that encourages the
reference motion to reach its strike phase at the specified time.

The reference joint positions are parameterized by phase as $\mathbf{q}^{\mathrm{ref}}(\phi)$.
The teacher outputs phase acceleration $\ddot\phi$, which is integrated to obtain phase velocity $\dot\phi$ and phase $\phi$.
Using the chain rule, $\dot{\mathbf{q}}^{\mathrm{ref}}=(\partial\mathbf{q}^{\mathrm{ref}}/\partial\phi)\dot\phi$, we constrain phase acceleration so that the updated phase rate keeps reference joint velocities within their prescribed limits.

Stage I thus yields a time-adaptive teacher that retimes the same reference motion to meet different strike deadlines.

Earlier feasible strike deadlines allow Stage II to match the same reference motion to faster incoming balls.

This expands the range of incoming-ball conditions that can be paired with the existing reference dataset for training.

\begin{table}[!t]
    \vspace*{2mm}
    \centering
    \footnotesize
    \setlength{\tabcolsep}{2pt}
    \renewcommand{\arraystretch}{1.12}
    \begin{tabular*}{\columnwidth}{@{\extracolsep{\fill}}lccc@{}}
        \toprule
        Term & Teacher & Critic & Actor\\
        \midrule
        Joint positions / velocities & $\checkmark$ & $\checkmark$ & $\checkmark$\\
        Previous action & $\checkmark$ & $\checkmark$ & $\checkmark$\\
        Body angular velocity & $\checkmark$ & $\checkmark$ & $\checkmark$\\
        Projected gravity & $\times$ & $\checkmark$ & $\checkmark$\\
        Root linear velocity & $\times$ & $\checkmark$ & $\times$\\
        World-frame root position & $\times$ & $\checkmark$ & $\checkmark$\\
        Tracked-body poses & $\times$ & $\checkmark$ & $\times$\\
        Reference joint state & $\checkmark$ & $\checkmark$ & $\times$\\
        Reference root pose & $\checkmark$ & $\checkmark$ & $\times$\\
        Strike goal (pos. / ori. / vel.) & $\checkmark$ & $\checkmark$ & $\times$\\
        Strike timing & $\checkmark$ & $\checkmark$ & $\times$\\
        Ball positions / velocities & $\times$ & $\checkmark$ & $\checkmark$\\
        Racket position / velocity & $\times$ & $\checkmark$ & $\checkmark$\\
        Initial heading direction & $\times$ & $\checkmark$ & $\checkmark$\\
        Landing-target XY & $\times$ & $\checkmark$ & $\checkmark$\\
        \midrule
        Additional information & $\phi$ & $\phi,\;\mathbf{z}^T$ & $\mathbf{z}^S$\\
        \bottomrule
    \end{tabular*}
    \caption{Policy observation}
    \label{tab:training_observations}
\end{table}

\setlength{\textfloatsep}{3pt}

\subsection{Stage II: Ball Trajectory Matching and Student Policy Distillation}

In Stage II, we sample ball launch positions and velocities, simulate trajectories with gravity,
drag, and ground rebound, and match trajectory points between the first and second bounces to
reference strike points.
A valid pair satisfies $\|\mathbf{b}_i(T_m)-\mathbf{c}_m\|_2\leq\epsilon$ and
$T_m\in[\bar T_m/\kappa,\bar T_m]$, where $\mathbf{b}_i$ is the incoming trajectory, $\mathbf{c}_m$
the reference contact point, and $\epsilon$ the matching tolerance.
The matched time becomes the teacher's deadline.
We use a library of 200 generated reference motions for training.

We load the frozen phase teacher and transfer its privileged skills to the student through
DAgger-style online distillation~\cite{ross2011dagger} within the teacher-guided PPO (TPPO) framework
(Fig.~\ref{fig:athlete_framework}\,\textcircled{3}).
In Stage II, PPO updates are disabled; only action and representation distillation objectives are
optimized.
During rollout collection, we gradually reduce the fraction of environments controlled by the
teacher and increase student control.
Teacher action labels are queried at every visited state, regardless of which policy executes the
actions.
This provides supervision on states reached after the student deviates from the teacher's
trajectories, mitigating the distribution shift associated with offline behavior cloning.
The teacher's joint actions already reflect its adaptation to the requested strike time.
We therefore distill only its 29-dimensional joint-action output, while the frozen teacher
continues to provide phase acceleration in the training environment to advance the reference
motion used by the teacher.

Instantaneous observations provide incomplete information about ball evolution and the forthcoming
striking motion.
We therefore construct a privileged training latent $\mathbf{z}_t^T$ and a deployable student latent
$\mathbf{z}_t^S$:
\begin{equation}
\begin{aligned}
\mathbf{z}_t^T&=E_T(\mathbf{h}_t^r,\mathcal{R}_t),\\
\mathbf{z}_t^S&=E_S(\mathbf{h}_t^r,\mathbf{h}_t^b),\\
\hat{\mathbf{a}}_t^S&=\mu_S(\mathbf{o}_t,\mathbf{z}_t^S).
\end{aligned}
\label{eq:teacher_student_intent}
\end{equation}

Here $\mathbf{h}_t^r$ and $\mathbf{h}_t^b$ contain robot-state and ball-position/velocity histories,
$\mathcal{R}_t$ is a future reference joint-position/velocity window, and $\mu_S$ is the student
action mean.
Both latents are 128-dimensional.
The student temporal encoder fuses robot and ball observations at five historical instants; the
privileged encoder uses robot-history queries to cross-attend to 12 future reference joint states.
The auxiliary reference encoder $E_T$ is learned in this stage and is distinct from the frozen phase
teacher $\pi_T$.
The student action network consumes $\mathbf{z}_t^S$ throughout, while reference motions remain
confined to the training supervision branch.

Three complementary losses supervise this transfer: Euclidean action distillation from the frozen
teacher, reconstruction of normalized future references from $\mathbf{z}_t^T$, and latent alignment
of $\mathbf{z}_t^S$ to a stop-gradient $\mathbf{z}_t^T$.
Let $\mathcal{H}$ be elementwise-averaged Smooth L1 loss, $D$ the reference decoder,
$\operatorname{sg}$ stop-gradient, and $\widetilde{\mathcal{R}}_t$ the normalized reference window:
\begin{equation}
\begin{aligned}
&\mathcal{L}_{\mathrm{act}}=\mathbb{E}\big[\|\hat{\mathbf{a}}_t^S-\mathbf{a}_t^T\|_2\big],\\
&\mathcal{L}_{\mathrm{ref}}=\mathbb{E}\big[\mathcal{H}(D(\mathbf{z}_t^T),\widetilde{\mathcal{R}}_t)\big],\\
&\mathcal{L}_{\mathrm{latent}}=\mathbb{E}\big[\mathcal{H}(\mathbf{z}_t^S,\operatorname{sg}(\mathbf{z}_t^T))\big],\\
&\mathcal{L}_{\mathrm{II}}=\mathcal{L}_{\mathrm{act}}+\mathcal{L}_{\mathrm{ref}}
+0.05\mathcal{L}_{\mathrm{latent}}.
\end{aligned}
\label{eq:joint_distillation}
\end{equation}
Expectations are taken over continually refreshed DAgger rollout batches, where
$\mathbf{a}_t^T$ denotes the teacher joint-action labels at the visited states.
Reference reconstruction gives the privileged latent explicit motion semantics, while latent
alignment transfers this supervision to a student representation conditioned only on observation
histories.
Consequently, $\mathbf{z}_t^S$ provides the policy with motion intent and timing information inferred
from history; neither reference motions nor explicit strike deadlines are required at execution.

Table~\ref{tab:training_observations} compares the frozen teacher, student actor, and Stage III
critic using the final fine-tuning configuration.
The asymmetric value network $V(\mathbf{o}_t^V,\operatorname{sg}(\mathbf{z}_t^T))$ receives
uncorrupted base observations, current references, and strike timing.
Value gradients do not update the privileged encoder.

\subsection{Stage III: Task Refinement via PPO}


We initialize from the distilled student, let it control all environments, and optimize tennis
performance with PPO using task (Fig.~\ref{fig:athlete_framework}\,\textcircled{4})..
To preserve the learned representations, both latent encoders and the reference decoder are frozen
while the policy and value heads are updated.

Task rewards guide pre-contact swinging and post-contact return quality:
$f_{\mathrm{swing}}$ encourages racket velocity along the reference striking direction near contact;
$f_{\mathrm{out}}$ rewards horizontal ball speed toward the landing target;
$f_{\mathrm{net}}$ rewards predicted net clearance; and
$f_{\mathrm{land}}$ rewards predicted landing accuracy for valid net-crossing trajectories.
The last three terms are evaluated from the post-strike ball state.

The total reward combines these task scores with auxiliary control rewards and penalties as
$r_t=\Delta t\sum_i w_i f_i(t)$, where $f_i(t)$ denotes the $i$th term and $w_i$ its weight,
with a control step of $\Delta t=0.02$\,s.
Episodes terminate when both pelvis and torso tilt exceed $50^\circ$.
A retained action-distillation coefficient of 0.05 regularizes refinement toward the teacher's
learned skills:
\vspace*{-5mm}
\begingroup

\begin{equation}
\mathcal{L}_{\mathrm{III}}
= \mathcal{L}_{\mathrm{PPO}}
+ 0.05\mathcal{L}_{\mathrm{act}}.
\label{eq:ppo_retained_distillation}
\end{equation}
\endgroup

Here $\mathcal{L}_{\mathrm{PPO}}$ includes the clipped policy surrogate, value loss, and entropy
regularization.
The 0.05 action-distillation weight is set independently of the Stage II latent-alignment weight;
reference reconstruction and latent losses are no longer optimized in Stage III.
The teacher supplies action supervision and reference-phase management only during simulation
training.
Deployment removes the teacher, privileged reference encoder, and decoder, retaining only $E_S$ and
the student action network.
This single policy produces 29-dimensional joint actions at 50\,Hz from current observations and
encoded histories, without reference motions or a phase controller.

During fine-tuning, we follow the ball-physics domain-randomization design of
LATENT~\cite{zhang2026latent}, adapting its parameterization and sampling ranges to our simulator to
address the sim-to-real dynamics gap.
At simulation initialization, ball mass, ball-ground restitution, ground tangential-speed retention,
and aerodynamic drag coefficients are sampled uniformly for each parallel environment.
We additionally randomize racket mass, center-of-mass offsets, and racket restitution, exposing the
student to variations in flight, ground rebound, and racket-contact dynamics during task refinement.
\vspace{-0.2cm}
\section{EXPERIMENTAL RESULTS}

\subsection{Reference Motion Coverage}
\label{sec:reference_motion_coverage}

We compare the original striking library with TGMM-generated reference libraries on 10,000 shared groups of forward-simulated incoming-ball trajectories, varying the number of qualified candidates per group, $K$, from 1 to 24.
Qualified trajectories clear the net and, before the second bounce, pass within 3\,m horizontally of the initial robot position at a ball-center height of at least 0.5\,m; filtering is independent of the motion library.
A match requires a point between the first and second bounces to lie within 0.20\,m of a reference strike point and satisfy the Stage I teacher's timing window, $\bar T_i/2\leq t+0.01\,\mathrm{s}\leq\bar T_i$, where $\bar T_i$ is the reference's nominal strike time.
A group succeeds if at least one of its first $K$ candidates matches.

\begin{figure}[!t]
    \vspace*{2mm}
    \centering
    \includegraphics[width=\columnwidth]{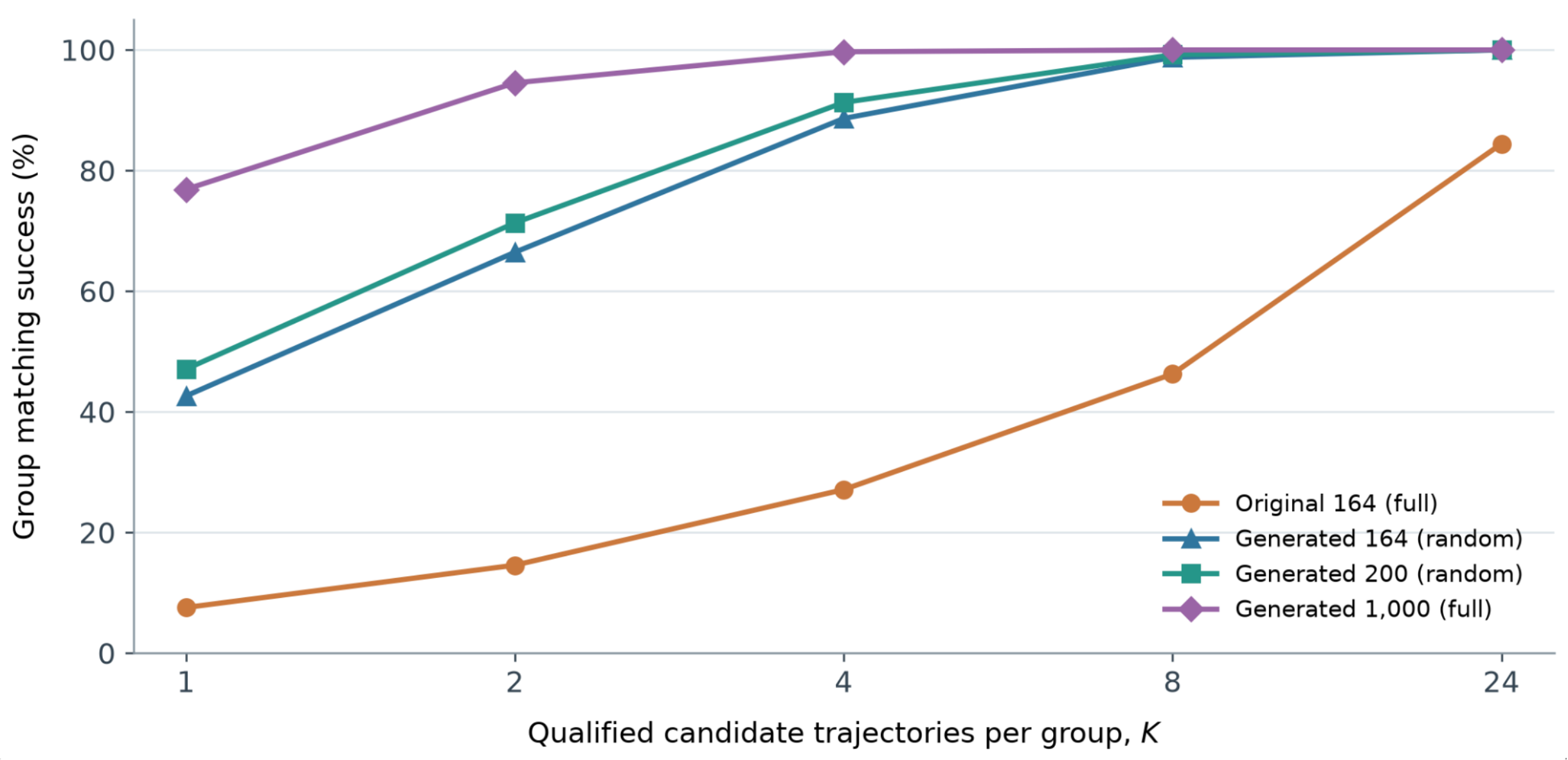}
    \caption{Spatiotemporal matching success rates of original and TGMM-generated reference libraries with varying numbers of qualified, forward-simulated ball trajectories per group, $K$.
    A group is successful if at least one candidate trajectory matches a reference motion.}
    \label{fig:matching_coverage_comparison}
    \vspace{-0.3cm}
\end{figure}

At the same library size of 164 motions and $K=1$, the generated references achieve 42.64\% matching success, compared with 7.57\% for the original library (Fig.~\ref{fig:matching_coverage_comparison}).
This improvement is consistent with the broader distribution of generated strike locations around the robot, whereas the original locations concentrate in front of it (Fig.~\ref{fig:strike_distribution}(b)).
TGMM thus expands spatiotemporal coverage even at a fixed library size.

Stage II uses 200 generated references to balance reference diversity and matching cost.
At this library size, the randomly sampled subsets reach 99.21\% mean group matching success at $K=8$, supporting broad coverage with a compact library and modest candidate budget.
Results for randomly sampled libraries are averaged over three subsets.

\begin{table*}[!t]
    \vspace*{1mm}
    \centering
    \footnotesize
    \setlength{\tabcolsep}{3pt}
    \renewcommand{\arraystretch}{1.18}
    \begin{tabular*}{\textwidth}{@{\extracolsep{\fill}}lcccccccc@{}}
        \toprule
        Policy & \multicolumn{2}{c}{Hit rate (\%) $\uparrow$} & \multicolumn{2}{c}{Return rate (\%) $\uparrow$} & \multicolumn{2}{c}{Landing err. (m) $\downarrow$} & \multicolumn{2}{c}{Sweet-spot dist. (cm) $\downarrow$}\\
        \cmidrule(lr){2-3}\cmidrule(lr){4-5}\cmidrule(lr){6-7}\cmidrule(lr){8-9}
        & Forehand & Backhand & Forehand & Backhand & Forehand & Backhand & Forehand & Backhand\\
        \midrule
        \mbox{w/o PPO fine-tuning} & 1.26 & 5.67 & 0.00 & 0.00 & 5.465 & 5.336 & 14.09 & 13.92\\
        \addlinespace[2pt]
        Full method & \textbf{94.97} & \textbf{95.04} & \textbf{91.19} & \textbf{92.20} & \textbf{1.150} & \textbf{1.314} & \textbf{5.61} & \textbf{6.54}\\
        \bottomrule
    \end{tabular*}
    \caption{Paired policy comparison on simulated incoming balls, with 159 forehand-side and 141 backhand-side trials per policy.
    Hit and return rates use all trials; return denotes net clearance.
    Landing error is the mean horizontal distance from the first post-strike landing to the target.
    Sweet-spot distance is the mean 3D distance from the ball center to the racket sweet spot at first contact.}
    \label{tab:simulation_results}
    \vspace{-0.3cm}
\end{table*}

\subsection{Simulation Experiments}
\label{sec:simulation_experiments}

\begin{figure}[!t]
    \vspace*{2mm}
    \centering
    \includegraphics[width=\columnwidth]{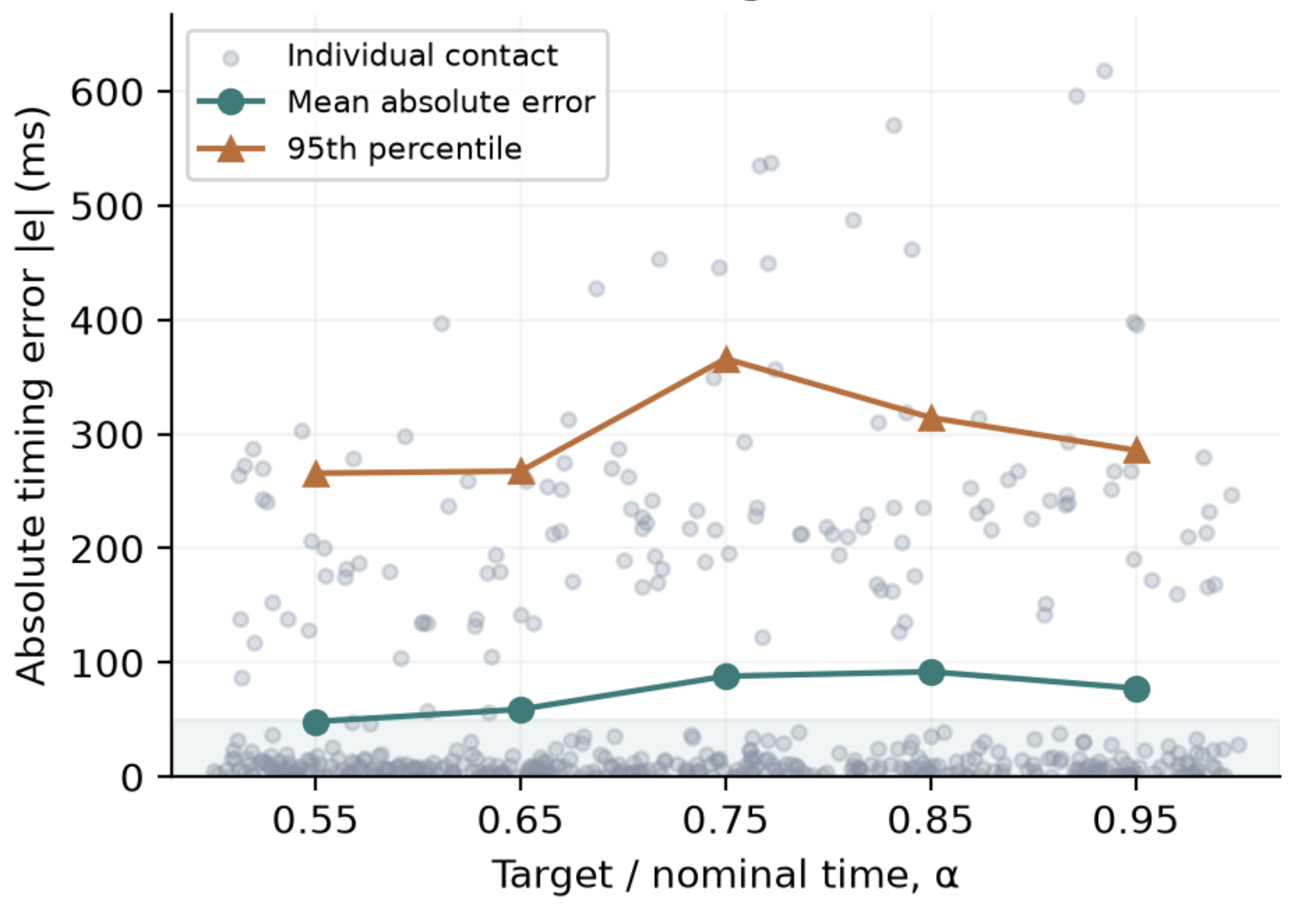}
    \vspace*{-4mm}
    \caption{Teacher contact timing across deadline ratios. Dots show absolute timing errors
        $|e|=|t_{\mathrm{contact}}-T_i|$ for 493 physical racket contacts, where $T_i$ and $t_{\mathrm{contact}}$ denote the target and actual contact times.
        Curves show binned mean absolute errors and 95th percentiles; shading marks errors within 50\,ms.}
    \label{fig:teacher_contact_timing}
\end{figure}

We evaluate the frozen teacher on 100 randomly selected motions from its training library under nominal physics, using gravity-compensated balls at stationary strike targets.
For each motion, we independently sample five deadlines $T_i=\alpha\bar T_i$, with $\alpha\sim\mathcal{U}(0.5,1)$.
During execution, the teacher-controlled phase updates progression through the reference motion.

Across 500 trials, the teacher makes physical racket contact in 493 trials (98.6\%), with 362 trials (72.4\%) achieving contact within $\pm50$\,ms of the target deadline.
Figure~\ref{fig:teacher_contact_timing} shows the absolute timing errors of the 493 contacts, with a mean absolute error of 71.23\,ms.
These results demonstrate that the teacher adapts whole-body motion timing to sampled deadlines, allowing the same reference to support more ball-arrival times and provide more motion--trajectory pairings for Stage II distillation.

We compare the Stage II distilled policy (w/o PPO fine-tuning) with the full method on 300 paired incoming balls and shared per-trial targets across three seeds (100 trials each).
Each policy uses its stage-specific deployment interface at 50\,Hz under nominal physics, resetting the robot and observation histories before each 8\,s trial.
Incoming balls use the deployment sampler without reference-motion matching, and landing targets follow the Stage III training distribution.

Grouping by the ball's side at its closest horizontal approach to the initial robot between the first two bounces yields 159 forehand-side and 141 backhand-side trials, independently of policy outcomes.
Table~\ref{tab:simulation_results} reports hit and return rates over all incoming balls, together with mean landing error and mean 3D ball-center-to-sweet-spot distance at first contact for struck balls.

Without PPO fine-tuning, forehand- and backhand-side hit rates are 1.26\% and 5.67\%, with no balls clearing the net; its detected contacts occur approximately 14\,cm from the racket sweet spot on average.
The full method raises hit rates to 94.97\%/95.04\% and return rates to 91.19\%/92.20\%, while reducing mean sweet-spot distances to 5.61/6.54\,cm and mean landing errors to 1.150/1.314\,m.
These results demonstrate improved contact precision and controlled returns across varying targets, supporting subsequent real-world deployment.
\begin{table}[!t]
    \centering
    \footnotesize
    \setlength{\tabcolsep}{2pt}
    \renewcommand{\arraystretch}{1.12}
    \begin{tabular*}{\columnwidth}{@{\extracolsep{\fill}}lcccc@{}}
        \toprule
        Method & \multicolumn{2}{c}{Hit rate (\%) $\uparrow$} & \multicolumn{2}{c}{Return rate (\%) $\uparrow$}\\
        \cmidrule(lr){2-3}\cmidrule(lr){4-5}
        & Forehand & Backhand & Forehand & Backhand\\
        \midrule
        LATENT~\cite{zhang2026latent} & --- & --- & 90.90 & 77.78\\
        AdaPT~\cite{huang2026adapt} & 52.00 & 50.67 & 21.33 & 19.99\\
        \midrule
        ATHLETE (ours) & \shortstack{$\mathbf{90.00}$\\$(27/30)$} & \shortstack{$\mathbf{86.67}$\\$(26/30)$} & \shortstack{$\mathbf{46.67}$\\$(14/30)$} & \shortstack{$\mathbf{53.33}$\\$(16/30)$}\\
        \bottomrule
    \end{tabular*}
    \caption{Real-robot results under each method's original evaluation protocol and metric definitions.
    AdaPT’s overall net-clearance rates are estimated by averaging the per-style products of hit and conditional net-clearance rates.
    Return denotes in-court return for LATENT and net clearance for AdaPT and ATHLETE. The motion-capture coverage areas are approximately
    $20\,\mathrm{m}\times8\,\mathrm{m}$ for AdaPT and
    $10.5\,\mathrm{m}\times5.5\,\mathrm{m}$ for ATHLETE. LATENT is evaluated on a full-size tennis court.}
    \label{tab:real_world_results}
\end{table}


\subsection{Real World Experiments and Benchmark}
\label{sec:real_world_experiments}

We evaluate single-shot striking and continuous human--robot rallies on a real humanoid robot
(Fig.~\ref{fig:athlete_framework}\,\textcircled{5}).
An external motion-capture system provides the ball's 3D position and the robot's root pose, and ball velocity is estimated from consecutive position observations.
Ball position and velocity are expressed in the robot's root frame according to the training observation definitions and supplied to the policy.

The motion-capture system covers a horizontal area of approximately $10.5\,\mathrm{m}\times5.5\,\mathrm{m}$, and we adapt the simulation training conditions to this confined space.
The reward for horizontal outgoing speed projected toward the landing target$ (5,0)$ saturates at $3.5$\,m/s, removing the incentive from this term for faster returns.
The policy runs at $50$\,Hz on a laptop equipped with an NVIDIA GeForce RTX 5070 Ti Laptop GPU, exchanging control commands and state information with the robot over wired Ethernet through the Unitree DDS interface.
The racket is attached to the robot's right hand using a 3D-printed connector.

\begin{figure}[!t]
    \vspace*{2mm}
    \centering
    \includegraphics[width=\columnwidth]{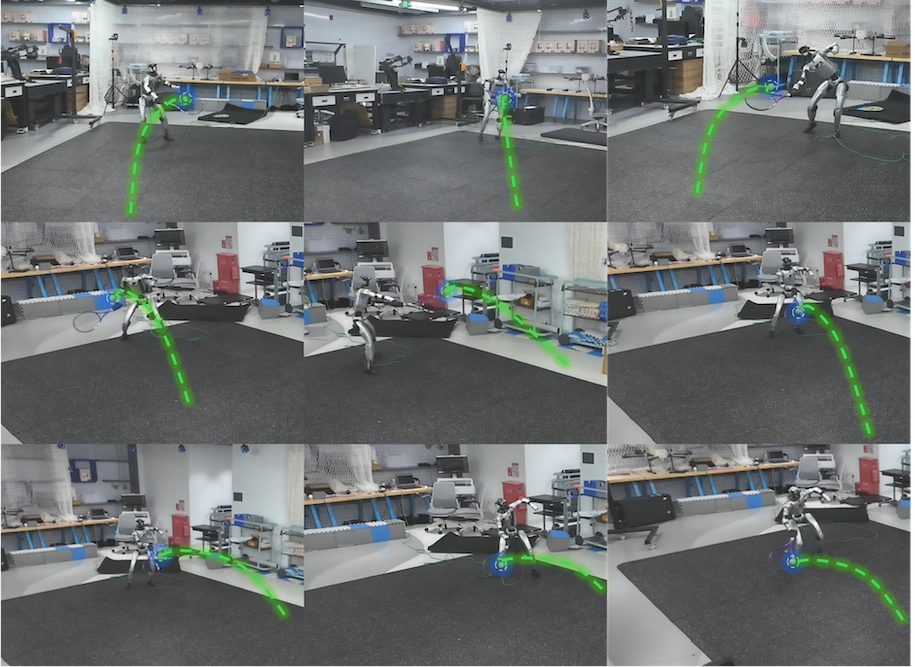}

    \caption{Real-world striking snapshots from the manual-throw experiment.
    }
    \label{fig:real_serve_back_experiment}
\end{figure}

The first experiment evaluates the robot's response to different incoming-ball locations using manual throws toward its forehand and backhand sides (Fig.~\ref{fig:real_serve_back_experiment}).
We conduct $30$ trials per side, for a total of $60$, varying the throw direction and location to cover different striking positions around the robot.
As shown in Table~\ref{tab:real_world_results}, the robot hits $27/30$ forehand balls ($90.00\%$) and $26/30$ backhand balls ($86.67\%$).
The robot clears the net in $14/30$ forehand trials ($46.67\%$) and $16/30$ backhand trials ($53.33\%$).
These results demonstrate reactive real-world striking across different incoming-ball locations.

Table~\ref{tab:real_world_results} includes published LATENT~\cite{zhang2026latent} human--robot rally results and AdaPT~\cite{huang2026adapt} results from 25 machine-fed trials per target region, retaining their original experiment environment.
Among the approaches using video-derived motions, ATHLETE achieves higher reported forehand and backhand hit and net-clearance rates than AdaPT,under their respective evaluation protocols. 

The second experiment evaluates continuous rallies between the robot and human partners with different levels of tennis experience (Fig.~\ref{fig:athlete_teaser}(e)--(h)).
The longest observed rallies are $5$ rounds with a novice player and $8$ rounds with a professional player.
These results demonstrate successful transfer to real hardware and sustained coordination of perception, locomotion, and striking during human--robot interaction.
\vspace*{-1mm}
\begin{table}[!t]
    \centering
    \footnotesize
    \setlength{\tabcolsep}{1pt}
    \renewcommand{\arraystretch}{1.18}
    \newcommand{\propertycell}[1]{\begin{tabular}[c]{@{}l@{}}#1\end{tabular}}
    \resizebox{\columnwidth}{!}{%
    \begin{tabular}{@{}lcccc@{}}
        \toprule
        \multicolumn{1}{c}{Property} & Vid2Player3D & LATENT & AdaPT & ATHLETE (ours)\\
        \midrule
        \propertycell{Existing public\\data reuse} & $\checkmark$ & --- & $\checkmark$ & $\checkmark$\\
        \addlinespace[4pt]
        \propertycell{No extra tennis\\MoCap demo\\collection} & $\checkmark$ & $\times$ & $\checkmark$ & $\checkmark$\\
        \addlinespace[4pt]
        \propertycell{Deployment\\with MoCap} & --- & $\checkmark$ & $\checkmark$ & $\checkmark$\\
        \addlinespace[4pt]
        \propertycell{No motion\\reference} & $\times$ & $\checkmark$ & $\times$ & $\checkmark$\\
        \addlinespace[4pt]
        \propertycell{No future-ball\\prediction} & $\times$ & $\checkmark$ & $\times$ & $\checkmark$\\
        \addlinespace[4pt]
        \propertycell{Single-level\\policy} & $\times$ & $\times$ & $\times$ & $\checkmark$\\
        \addlinespace[4pt]
        \midrule
        Training GPUs & $1\times$\,A100 & $8\times$\,--- & $4\times$\,RTX~4090 & $1\times$\,RTX~4090\\
        \bottomrule
    \end{tabular}%
    }
    \caption{Qualitative comparison of reported configurations. Dashes indicate unavailable or inapplicable entries.
    }
    \label{tab:qualitative_comparison}
\end{table}

\subsection{Qualitative Comparison}
\label{sec:qualitative_comparison}

Table~\ref{tab:qualitative_comparison} compares the reported data sources, control architectures, and training configurations of Vid2Player3D~\cite{zhang2023vid2player3d}, LATENT, AdaPT, and ATHLETE.
LATENT reports strong in-court return performance using five hours of newly captured motion primitives from amateur players and eight training GPUs.
ATHLETE avoids dedicated tennis motion-capture collection by reusing public locomotion and sparse video-derived strikes, and trains its policies on a single RTX 4090. AdaPT also uses video-derived motions for quantitative evaluation, but retains explicit motion planning, future-ball prediction, and reference tracking at deployment.
ATHLETE converts composed references into a single reactive policy that supports forehand and backhand returns and continuous rallies without these online components.
Together with the real-world results, this demonstrates the value of reusing motion data while simplifying deployment.
All three real-robot evaluations currently use external motion capture for state estimation.

\vspace{-0.3cm}
\section{Conclusion}

In this work, inspired by how humans acquire complex skills by combining reusable motor experience with task-specific demonstrations, we presented ATHLETE, a framework for learning reactive humanoid tennis skills from generic locomotion data and ordinary tennis videos, without dedicated tennis motion-capture demonstrations. By combining trajectory-guided motion generation with teacher--student learning, ATHLETE learns a deployable policy that responds to incoming balls without reference-motion inputs at deployment. Experiments on a real humanoid robot demonstrate forehand and backhand returns as well as continuous rallies with human players. The current system relies on external motion capture for ball-state estimation. Future work will investigate onboard ball perception and preservation of motion-style diversity.






\vspace*{-2mm}

\end{document}